%% file: main.tex
\documentclass[sigconf]{acmart}
\usepackage{booktabs}
\usepackage{amsmath}
\usepackage{array}
\usepackage{microtype}
\usepackage{enumitem}

\AtBeginDocument{%
  }

\copyrightyear{2026}
\acmYear{2026}
\setcopyright{cc}
\setcctype{by-nc-nd}
\acmConference[UbiComp Companion '26]{Companion of the 2026 ACM International Joint Conference on Pervasive and Ubiquitous Computing}{October 11--15, 2026}{Shanghai, China}
\acmBooktitle{Companion of the 2026 ACM International Joint Conference on Pervasive and Ubiquitous Computing (UbiComp Companion '26), October 11--15, 2026, Shanghai, China}
\acmDOI{10.1145/3798063.3841770}
\acmISBN{979-8-4007-2533-3/2026/10}

\begin{document}

\title[ReliaGate: Reliability Routing]{ReliaGate: Reliability Routing for Low-Stakes Wearable Stress Prediction}


\author{Jaden Moon}
\orcid{0009-0001-0251-6102}
\affiliation{%
  \institution{Dartmouth College}
  \city{Hanover}
  \state{New Hampshire}
  \country{USA}
}
\email{jaden.moon.27@dartmouth.edu}

\author{Yu Wu}
\orcid{0009-0004-3709-7472}
\affiliation{%
  \institution{Dartmouth College}
  \city{Hanover}
  \state{New Hampshire}
  \country{USA}
}
\email{yvonne.wu@dartmouth.edu}

\author{Arvind Pillai}
\orcid{0000-0002-2489-1130}
\affiliation{%
  \institution{Dartmouth College}
  \city{Hanover}
  \state{New Hampshire}
  \country{USA}
}
\email{arvind.thanga.chellappa.pillai.gr@dartmouth.edu}

\author{Andrew Campbell}
\orcid{0000-0001-7394-7682}
\affiliation{%
  \institution{Dartmouth College}
  \city{Hanover}
  \state{New Hampshire}
  \country{USA}
}
\email{andrew.t.p.campbell@gmail.com}

\renewcommand{\shortauthors}{Moon et al.}

\begin{abstract}
\input{sections/00_abs}
\end{abstract}




\maketitle

\input{sections/01_intro}
\input{sections/02_related_work}
\input{sections/03_method}
\input{sections/04_evaluation}
\input{sections/05_results}
\input{sections/06_discussion}
\input{sections/07_conclusion}

\begin{acks}
We thank the anonymous reviewers for their constructive feedback. We also thank
the creators and maintainers of the WESAD, UBFC-Phys, EmpaticaE4Stress, and
PhysioNetE4 datasets for making these resources available to the research
community.
\end{acks}

\clearpage
\bibliographystyle{ACM-Reference-Format}
\bibliography{references}

\end{document}

%% file: sections/00_abs.tex
We study when a wearable stress system should surface a prediction rather than
change it. In low-stakes reflection and summary settings, aggregate accuracy is
insufficient because withholding can reduce error while leaving some people with
little or no information. We formulate fixed-label reliability routing: after a
locked classifier emits a protocol-defined stress/non-stress label, a post-hoc
gate surfaces that unchanged label or withholds it as unavailable. ReliaGate assembles established confidence, signal-quality/trust, agreement,
train-standardized atypicality, and train-fitted geometry cues into a post-hoc
correctness score. We evaluate four wearable datasets using subject-disjoint
folds, validation-selected routing, paired held-out-subject intervals, and
pooled and per-subject analyses. WESAD point estimates favored ReliaGate, UBFC-Phys primary coverage/risk
intervals favored ReliaGate, and E4 checks were mixed. ReliaGate provides an
operational framework for studying surfaced-label error, output availability,
and accepted-output distribution across subjects, without revising labels or
providing clinical or finite-sample risk guarantees.

%% file: sections/01_intro.tex
\section{Introduction}
\label{sec:introduction}

Wearable stress inference is increasingly positioned as support for low-stakes
reflection and summaries rather than diagnosis. Self-tracking and
personal-informatics research frame such systems as tools for feedback and
reflection~\cite{feng2021selftracking,li2010stageBasedPersonalInformatics}.
Wearable stress studies commonly use photoplethysmography (PPG)-derived blood
volume pulse (BVP), electrodermal activity (EDA), and skin temperature (TEMP),
with protocol stressor and rest periods providing stress/non-stress reference
labels~\cite{pinge2024wearables}. Thus, the binary labels studied here are
protocol-defined and non-clinical. Even in this low-stakes setting, however, a
surfaced label can shape user interpretation: a scoping review of
health-monitoring wearables reports that false positives may cause anxiety or
distress, whereas false negatives may provide unfounded reassurance
~\cite{capulli2025ethicalLegalWearables}.

The central problem is not only whether a stress classifier is accurate on
average, but whether a particular output is reliable enough to show.
Wearable PPG and ambulatory EDA are vulnerable to artifacts
~\cite{lee2020ppgMotionArtifact,gashi2020edaArtifacts}, participant variation
in WESAD and protocol differences across studies can challenge generalized
performance~\cite{li2024personalizedWearables,mishra2020reproducibility}, and
confidence calibration can degrade under dataset shift
~\cite{ovadia2019trust}. Aggregate accuracy does not resolve these
per-output decisions. We therefore ask: \emph{when should a wearable
well-being system surface a stress/non-stress label at all?}

We study this as \emph{reliability routing}, an application-level use of
selective classification~\cite{chow1970optimum,elYaniv2010foundations,
geifman2017selective}. After a locked classifier emits a label, a post-hoc
gate either surfaces that unchanged output or withholds it as unavailable.
Coverage is the fraction surfaced, and accepted risk is the error rate among
surfaced labels. These quantities must be read together because low accepted
risk can result from withholding most predictions
~\cite{traub2024selectiveEvaluation}. Routing changes availability, not the
upstream model or predicted class.

We introduce \textbf{ReliaGate}, a correctness scorer that combines
inference-time confidence, signal quality and modality trust, cross-modal
agreement, train-standardized atypicality, and train-fitted embedding geometry.
We evaluate four wearable datasets using subject-disjoint folds with fold-local
scorer and threshold selection. Beyond pooled error, we test whether routing
leaves held-out subjects with little or no output, since selective systems can
appear more reliable by withholding unevenly across people
~\cite{traub2024selectiveEvaluation,jones2021selectiveDisparities,
lee2021fairSelective}.

\clearpage

\noindent\textbf{Contributions.} This paper contributes:
\begin{itemize}[leftmargin=*,nosep]

\item A fixed-label \emph{reliability routing} formulation that separates
stress-label prediction from the decision to surface or withhold that label.

\item \textbf{ReliaGate}, a leakage-audited assembly of established confidence,
signal-quality/trust, agreement, atypicality, and train-fitted geometry cues
without modifying the upstream classifier.

\item A subject-disjoint four-dataset evaluation with fold-local selection,
paired held-out-subject intervals, and a per-subject availability lens based on
Reach and coverage alongside surfaced-label error.

\end{itemize}

%% file: sections/02_related_work.tex
\section{Related Work}
\label{sec:related-work}

\noindent\textbf{Wearable stress sensing and reliability.}
Wearable stress studies use PPG-derived BVP, EDA, skin temperature, and
activity signals~\cite{pinge2024wearables, schmidt2018wesad}. The challenge for
well-being systems is not only predicting a stress label, but deciding whether
a particular label is reliable enough to show. Stress-classifier performance
can vary across participants, sensors, protocols, and stressors
~\cite{li2024personalizedWearables,mishra2020reproducibility,
prajod2024stressorTypeMatters}. Signal quality adds another source of
variation: wearable PPG is vulnerable to motion artifacts
~\cite{lee2020ppgMotionArtifact}, and ambulatory EDA to recording, movement,
and environmental artifacts~\cite{taylor2015edaArtifacts,
gashi2020edaArtifacts}. More generally, confidence calibration can degrade
under dataset shift~\cite{ovadia2019trust}. Together, these findings motivate
a per-output surfacing decision rather than reliance on aggregate accuracy
alone~\cite{traub2024selectiveEvaluation}.

\noindent\textbf{Post-hoc reliability estimation and selective routing.}
Prior reliability methods address different parts of this decision. Calibration
maps scores to probabilities intended to reflect correctness likelihood
~\cite{guo2017calibration}. Maximum softmax probability supports
confidence-only error screening~\cite{hendrycks2017baseline}, whereas failure
prediction learns scores that separate correct from erroneous outputs of an
existing classifier~\cite{corbiere2019failurePrediction}.
Representation-based out-of-distribution methods provide Mahalanobis and
nearest-neighbor atypicality scores
~\cite{lee2018mahalanobis,sun2022knn}, and trusted multi-view classification
uses view-specific evidence and uncertainty during fusion
~\cite{han2021trustedMultiview}.

ReliaGate assembles these established cues for fixed-label surfacing: after the
wearable label is fixed, it scores likely correctness and surfaces or withholds
that unchanged label. Reject-option and selective-classification work established
error--rejection and risk--coverage formulations
~\cite{chow1970optimum,elYaniv2010foundations,geifman2017selective}. Our
contribution is the leakage-audited integration of the cues with
validation-locked routing and a per-subject availability lens, not a new
calibration method, distance metric, OOD detector, abstention theory, or
finite-sample risk guarantee~\cite{angelopoulos2025learnThenTest}. Per-subject
diagnostics are not a demographic fairness analysis
~\cite{jones2021selectiveDisparities,lee2021fairSelective}.

%% file: sections/03_method.tex
\section{ReliaGate: Representation-Augmented Reliability Routing}
\label{sec:method}

\begin{figure*}[t]
  \centering
  \includegraphics[width=0.96\textwidth]{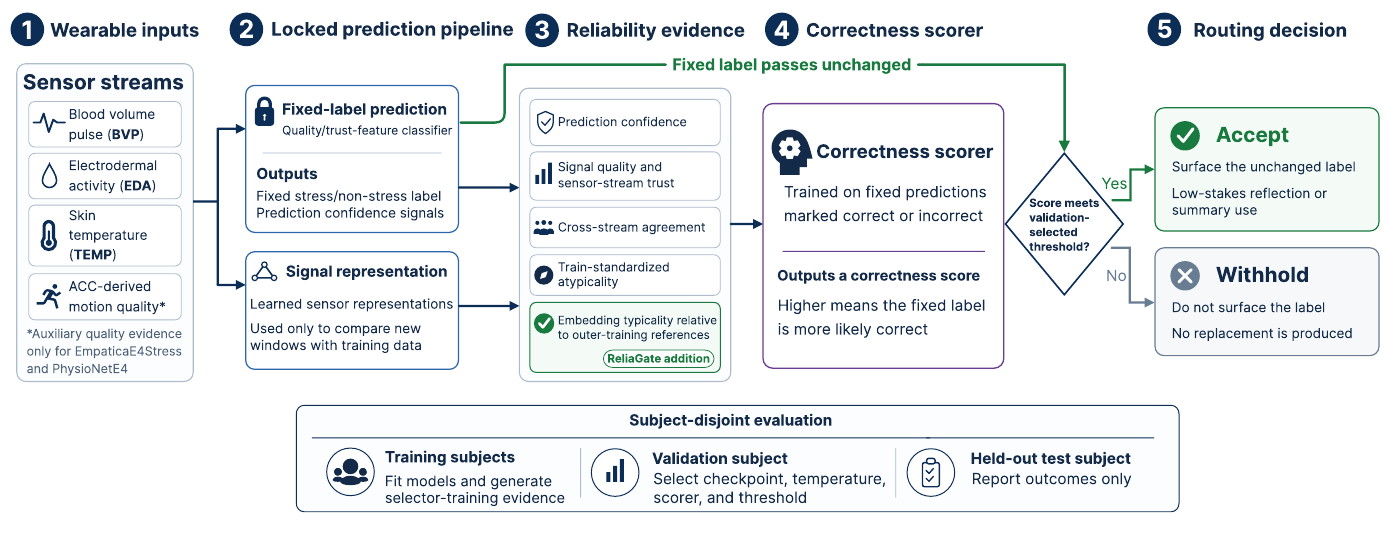}
\caption{ReliaGate routes an unchanged label using confidence, quality/trust,
agreement, train-standardized atypicality, and train-fitted geometry.
Outer-training subjects fit models and references; validation selects
checkpoint, temperature, scorer, and threshold; test data are reporting-only.
Canonical feature-model predictions are subject-wise out of fold (OOF), while
primary geometry may retain same-subject windows.}
\Description{A left-to-right five-stage pipeline. Wearable BVP, EDA, TEMP,
and optional ACC enter a locked predictor that emits a fixed stress/non-stress
label and learned representations. Reliability evidence combines prediction
confidence, signal quality and modality trust, cross-stream agreement,
train-standardized atypicality, and embedding geometry relative to
outer-training references. A correctness scorer assigns a score, and a
validation-selected threshold either surfaces the unchanged label or withholds
it; no replacement label is produced. A lower panel shows subject-disjoint
roles: training fits models and references, validation selects the checkpoint,
temperature, scorer, and threshold, and held-out test subjects are used only
for reporting.}
  \label{fig:ReliaGate-pipeline}
\end{figure*}

ReliaGate implements the surfacing decision posed in
Section~\ref{sec:introduction}. In each subject-disjoint outer fold, the
upstream pipeline is trained and locked before routing. ReliaGate scores the
fixed label from inference-time evidence and surfaces it unchanged or withholds
it as unavailable. Figure~\ref{fig:ReliaGate-pipeline} summarizes the pipeline
and data roles.

Let $D$ denote a dataset, $f$ an outer fold, and $x_i$ window $i$. The locked
canonical feature model emits logit $\ell_{i,f}$, stress probability
$\rho_{i,f}=\sigma(\ell_{i,f})$, where $\sigma$ is the logistic sigmoid, and
fixed label $\hat y_{i,f}=\mathbf{1}\{\rho_{i,f}\geq1/2\}$, where
$\mathbf{1}\{\cdot\}$ is the indicator function. The no-geometry system (NG)
uses reliability vector $\phi_{i,f}^{\mathrm{NG}}$. ReliaGate (RG) appends geometry vector $\gamma_{i,f}$, with $\oplus$
denoting concatenation:
$\phi_{i,f}^{\mathrm{RG}}=\phi_{i,f}^{\mathrm{NG}}\oplus\gamma_{i,f}$. For system
$r\in\{\mathrm{NG},\mathrm{RG}\}$, correctness scorer $g_f^{(r)}$ produces
\[
s_{i,f}^{(r)}=g_f^{(r)}\!\left(\phi_{i,f}^{(r)}\right).
\]
Larger $s_{i,f}^{(r)}$ means greater predicted correctness. A fold-specific
threshold $\tau_f^{(r)}$ accepts the unchanged label when
$s_{i,f}^{(r)}\geq\tau_f^{(r)}$ and withholds it otherwise. We use the score
for ranking rather than treating it as a guaranteed calibrated correctness
probability.

\noindent\textbf{Running example.}
Consider a held-out WESAD window for which the locked model predicts stress.
ReliaGate does not reconsider that class: it combines confidence, BVP/EDA/TEMP
quality and agreement, and distances to outer-training representations; maps
these cues to $s_{i,f}^{(\mathrm{RG})}$; and surfaces the stress label only when
that score meets the validation-selected threshold $\tau_f^{(\mathrm{RG})}$.
The reference label is used only afterward for evaluation.

\subsection{Upstream Physiological Predictors}
\label{sec:upstream}

The upstream predictors provide fixed labels and modality representations. Each
fold uses modality-specific temporal convolutional networks (TCNs) for BVP,
EDA, and, where available, TEMP, adapted from the dilated residual TCN family
~\cite{bai2018tcn}. The executed non-causal networks use five residual blocks
with dilations $1,2,4,8,16$, adaptive pooling, and a 64-dimensional projected
representation. BVP uses two stride-4 stem convolutions; EDA and TEMP use
stride 1. Training uses class-weighted cross-entropy and AdamW for at most 100
epochs, with validation macro-F1 checkpointing and patience 15. Training-window
median/interquartile-range scaling is protected against zero scale and clipped
to $[-10,10]$. Temperature scaling is fitted on the same validation subject,
bounded to $[0.05,20]$, and leaves the modality argmax unchanged
~\cite{guo2017calibration}.

ReliaGate separates three per-modality evidence types before scoring
correctness: posterior decisiveness, agreement with other streams, and raw
signal quality. Let $\mathcal M_D$ be dataset $D$'s set of primary modalities;
every analyzed dataset has at least two. For $m\in\mathcal M_D$, let
$p_{i,f}^{m}$ be modality $m$'s temperature-scaled stress posterior and let
$\bar p_{i,f}^{-m}$ be the arithmetic mean of the remaining modalities'
posteriors. After clipping probabilities away from 0 and 1 and using natural
logarithms, define binary entropy
$H_b(p)=-p\log p-(1-p)\log(1-p)$ and normalized confidence
$c_b(p)=1-H_b(p)/\log 2$. Let $\mathrm{JS}_b$ be the equal-weight
Jensen--Shannon divergence between Bernoulli posteriors $(1-p,p)$ and
$(1-q,q)$~\cite{lin1991divergence}. We use
\[
\begin{aligned}
Q_{\mathrm{conf},i,f}^{m}&=c_b(p_{i,f}^{m}),\\
Q_{\mathrm{cons},i,f}^{m}&=
\left[1-\frac{\mathrm{JS}_b(p_{i,f}^{m},\bar p_{i,f}^{-m})}{\log 2}\right]
c_b(\bar p_{i,f}^{-m}).
\end{aligned}
\]
$Q_{\mathrm{conf}}^{m}$ measures decisiveness. $Q_{\mathrm{cons}}^{m}$ rewards
agreement only when the other modalities are also decisive. Its second factor
is our heuristic, motivated by sample-dependent view reliability
~\cite{han2021trustedMultiview}.

For modality $m$, index its $J_m$
signal checks by $j$. Each mapped component $q_{i,j,f}^{m}\in[0,1]$ is fitted
using outer-training data, and raw quality is its stabilized geometric mean:
\[
Q_{\mathrm{raw},i,f}^{m}=
\sqrt[J_m]{\prod_{j=1}^{J_m}\max\{q_{i,j,f}^{m},10^{-6}\}}.
\]
WESAD checks finite values, flatlines/range, BVP pulse behavior, EDA
spikes/derivatives, and TEMP repeated values/derivatives. Studies of wearable
photoplethysmography and ambulatory EDA motivate these families
~\cite{lee2020ppgMotionArtifact,taylor2015edaArtifacts,
gashi2020edaArtifacts}; the mappings are manuscript-specific. With locked
scale $\beta_f$ and coefficients $\lambda_{r,f},\lambda_{c,f},\lambda_{s,f}$
on raw quality, confidence, and consistency, a softmax over modalities forms
weights summing to one:
\[
w_{i,f}^{m}=\operatorname{softmax}_{m}\!\left\{
\beta_f(\lambda_{r,f}Q_{\mathrm{raw},i,f}^{m}+
\lambda_{c,f}Q_{\mathrm{conf},i,f}^{m}+
\lambda_{s,f}Q_{\mathrm{cons},i,f}^{m})\right\}.
\]
WESAD fixes $\beta_f=1$ and
$\lambda_{r,f}=\lambda_{c,f}=\lambda_{s,f}=1/3$. Other datasets use locked coefficients fixed before held-out-test reporting. Accelerometry (ACC) is
auxiliary motion-quality evidence only in the two E4 datasets.

The executed canonical model is intentionally treated as a locked substrate for
routing. For WESAD, it is a median-imputed, standardized, class-balanced
$\ell_2$ logistic regression (inverse regularization $C=1$) on the 12
$[Q_{\mathrm{raw}}^{m},Q_{\mathrm{conf}}^{m},Q_{\mathrm{cons}}^{m},w^{m}]$
entries for BVP, EDA, and TEMP. UBFC-Phys uses the corresponding eight BVP/EDA
entries. Each E4 contract uses 12 primary-stream entries plus three
ACC-quality entries. These contracts contain no signed posterior or logit, so
they are quality/trust-feature classifiers rather than directional posterior
fusion.

For selector training, only the final canonical logistic model is refitted
without each outer-training subject and scores that subject's precomputed rows,
following stacked generalization~\cite{wolpert1992stacked}. Modality networks,
temperatures, quality mappings, and trust rows are reused. The canonical
feature-model prediction is subject-wise out of fold (OOF), whereas the
complete upstream stack is not.

\subsection{Correctness-Scoring Objective}
\label{sec:correctness}

ReliaGate learns correctness, not stress class. Let $y_i\in\{0,1\}$ be the
protocol-derived reference label. The binary correctness target is
\begin{equation}
z_{i,f}=\mathbf{1}\{\hat y_{i,f}=y_i\}.
\label{eq:correctness-target}
\end{equation}
Maximum predicted-class probability is the confidence-only baseline
~\cite{hendrycks2017baseline}. ReliaGate learns a post-hoc score for
$z_{i,f}$~\cite{corbiere2019failurePrediction} and thresholds it within
reject-option and selective-classification frameworks~\cite{chow1970optimum,
elYaniv2010foundations,geifman2017selective}. Neither $y_i$ nor $z_{i,f}$ is
available at inference time.

\subsection{Non-Geometry Reliability Evidence}
\label{sec:nongeometry}

The non-geometry vector captures what can be known about a fixed output without
using embedding geometry. In WESAD it has 43 features: seven deterministic
descriptors of $(\rho_{i,f},\ell_{i,f})$, 12 per-modality quality/trust values,
21 cross-modality summaries, and three atypicality norms. The ordered manifest fixes the column order; the summaries comprise five
statistics for each $Q$ family and the weights, plus weight entropy.

Let $b_{i,f}\in\mathbb R^{40}$ contain the first 40 features and
$\widetilde b_{i,f}$ its median-imputed form. Training-row means $\mu_f$ and
standard deviations $\sigma_f$ define
$v_{i,f}=(\widetilde b_{i,f}-\mu_f)\oslash\sigma_f$, where $\oslash$ is
elementwise division and zero-variance features use unit scale. We append its
Euclidean, maximum-absolute, and mean-absolute magnitudes:
\[
\delta_{i,f}=
\left[
\|v_{i,f}\|_2,\,
\|v_{i,f}\|_\infty,\,
40^{-1}\|v_{i,f}\|_1
\right].
\]
These are heuristic train-standardized atypicality cues, not formal shift
estimates. The citation motivates only the broader concern that uncertainty can
degrade under shift~\cite{ovadia2019trust}. Transforms are applied unchanged
to validation and test rows. Dimensions are 35 for UBFC-Phys and 51 for each
E4 dataset. Manifests exclude labels, targets, losses, identifiers, thresholds,
and routing outcomes, consistent with leakage
guidance~\cite{kapoor2023leakage}.

\subsection{Train-Fitted Representation Geometry}
\label{sec:geometry}

ReliaGate adds embedding evidence because an output may be less reliable when
its learned representation is atypical relative to training examples. Geometry
is computed in each modality's locked outer-fold representation space. WESAD
retains embedding norm, centroid distance, diagonal Mahalanobis distance with
variance floor $10^{-6}$, Ledoit--Wolf Mahalanobis distance, and mean/minimum
five-neighbor distances~\cite{lee2018mahalanobis,
ledoit2004wellconditioned,sun2022knn}. These sources motivate the distance
families; ReliaGate uses them as correctness features, not an
out-of-distribution detector.

Six families across three modalities yield 18 WESAD features. Their mean,
sample standard deviation, minimum, maximum, and range add 30 more, giving 48
geometry features and $43+48=91$ combined features. UBFC-Phys and E4 use
standardized embeddings and locked five-family contracts, yielding 35 and 40
geometry features, respectively.

Validation and test geometry use only outer-training references. For
selector-training rows, other windows from the query subject remain in the
reference pool, only the exact self-neighbor is removed, and the encoder may
have seen that subject. Thus primary geometry is not subject-excluded at either
level. A WESAD sensitivity removes the query subject from the references but
retains the encoder; it tests reference sharing, not subject-excluded encoder
training.

\subsection{Selector Training and Leakage Controls}
\label{sec:scorers}

The selector turns reliability evidence into the routing score. Within each
outer fold, NG and RG train logistic-regression and histogram
gradient-boosted decision-tree (GBDT) correctness candidates. Locked feature
orders, constructors, seeds, and fold selections are recorded, and validation
selects the family. WESAD ReliaGate minimizes the sum of within-fold ranks
across four validation criteria: AURC and a manuscript-defined generalized-risk
area that combines accepted errors with a fixed withholding penalty (lower is
better), and correctness AUROC and AUPRC (higher is better)
~\cite{traub2024selectiveEvaluation}. AUROC and AUPRC denote areas under the
receiver-operating-characteristic and precision--recall curves. Other feature
sets use AURC, generalized-risk area, AUROC, a simpler-model rank, and
candidate identifier in a locked lexicographic order. The WESAD ReliaGate GBDT
also reserves 15\% of selector-training rows for row-level early stopping, not
a subject holdout. Selection therefore has two stages: validation first chooses
the scorer family, then chooses that scorer's operating threshold.

For threshold selection, suppress the system superscript. Let $V_f$ be fold
$f$'s validation windows, $t$ a candidate threshold, and
$A_{v,f}(t)=\{i\in V_f:s_{i,f}\geq t\}$ the accepted set. Coverage is
$|A_{v,f}(t)|/|V_f|$, and risk is the fraction of its windows with
$z_{i,f}=0$. At the primary empirical criterion $\alpha=0.10$, every unique
validation score is considered. We maximize nonempty validation coverage
subject to risk at most $\alpha$, breaking ties by lower risk and then higher
threshold. If no threshold is feasible, the fold withholds every test
prediction. The selected scorer and threshold are applied unchanged to the
held-out test subject.

The same validation subject selects the upstream checkpoint, temperature,
scorer family, and threshold. Held-out test subjects remain isolated from
fitting and selection, but overlapping validation windows can make selection
optimistic. Thus $\alpha$ is an empirical criterion, not a finite-sample
guarantee or safety parameter. Because NG and RG may select different scorer
families, the primary comparison is system-level; the frozen
43-versus-91-feature GBDT analysis is a post-hoc, scorer-matched sensitivity,
not a causal ablation.

%% file: sections/04_evaluation.tex
\section{Task, Datasets, and Evaluation Protocol}
\label{sec:evaluation-protocol}

\subsection{Datasets}

The four datasets provide complementary within-dataset evaluations rather than
replications or transfer tests. WESAD supports primary mechanism and robustness
analyses~\cite{schmidt2018wesad}; UBFC-Phys provides an additional controlled
BVP/EDA evaluation~\cite{meziati2021ubfc,ubfcpaper}; and the E4 datasets
provide protocol checks~\cite{campanella2024databrief,
hongn2025scientificdata}. Each dataset is trained and evaluated independently.
Because each outer fold holds out one test subject, folds equal analyzed
subjects.

\begin{table*}[!t]
\caption{Analyzed post-filter samples, manuscript-defined binary tasks, and
held-out-subject folds.}
\label{tab:datasets}
\centering
\footnotesize
\setlength{\tabcolsep}{3.2pt}
\renewcommand{\arraystretch}{1.08}
\begin{tabular}{@{}
  >{\raggedright\arraybackslash}p{2.80cm}
  >{\raggedright\arraybackslash}p{2.45cm}
  >{\raggedright\arraybackslash}p{2.05cm}
  >{\raggedright\arraybackslash}p{2.20cm}
  >{\raggedright\arraybackslash}p{2.15cm}
  >{\raggedleft\arraybackslash}p{0.70cm}
  >{\raggedleft\arraybackslash}p{0.95cm}
  >{\raggedleft\arraybackslash}p{1.35cm}
@{}}
\toprule
Dataset
& Evaluation role
& Model inputs
& Non-stress periods
& Stress periods
& Folds
& Windows
& Counts N/S \\
\midrule

WESAD~\cite{schmidt2018wesad}
& Primary mechanism and robustness analysis
& BVP, EDA, TEMP
& Baseline; amusement
& TSST
& 15
& 2,986
& 2,099 / 887 \\

UBFC-Phys~\cite{meziati2021ubfc,ubfcpaper}
& Controlled BVP/EDA evaluation
& BVP, EDA
& T1 rest
& T2 speech; \newline T3 arithmetic
& 56
& 1,848
& 616 / 1,232 \\

EmpaticaE4Stress~\cite{campanella2023empaticae4stress,
campanella2024databrief}
& E4 protocol check
& BVP, EDA, TEMP; ACC quality
& Rest blocks
& Task blocks
& 29
& 4,301
& 1,044 / 3,257 \\

PhysioNetE4~\cite{hongn2025physionet,hongn2025scientificdata}
& Acute-stressor protocol check
& BVP, EDA, TEMP; ACC quality
& Baseline/rest blocks
& Stressor blocks
& 34
& 4,784
& 3,973 / 811 \\

\bottomrule
\end{tabular}

\vspace{0.15em}
\begin{minipage}{0.96\textwidth}
\footnotesize
\emph{Note.} Counts are post-filter manifest windows; N/S means
non-stress/stress. TEMP, ACC, and TSST denote skin temperature, accelerometry,
and Trier Social Stress Test. ACC is auxiliary where listed; labels
operationalize protocol periods and are not diagnoses.
\end{minipage}
\end{table*}

Source protocols and manuscript choices are summarized here. WESAD maps
baseline/amusement to non-stress and Trier Social Stress Test (TSST) to stress, excludes other codes, and
uses within-label 60~s windows with a 10~s stride after 10~s trims
~\cite{schmidt2018wesad}; ACC is unused. UBFC-Phys maps T1 to non-stress and
T2--T3 to stress and windows each task at 60~s/10~s after 10~s trims
~\cite{meziati2021ubfc,ubfcpaper}. EmpaticaE4Stress maps rest/task blocks to
non-stress/stress and uses 60~s/10~s windows; boundary trims apply only when a
full 60~s window remains~\cite{campanella2023stress,
campanella2024databrief}. PhysioNetE4 maps STRESS-session baseline/rest and
induced-stressor tags to non-stress/stress, excludes exercise, and uses
within-tag 30~s/10~s windows~\cite{hongn2025physionet,
hongn2025scientificdata}. Subjects f07 and f14 are excluded, and duplicate S02
tails are truncated under repository constraints
~\cite{hongn2025physionetConstraints}. ACC is quality-only for both E4 datasets.

\subsection{Compared Decision Systems}

All systems act on the same fixed label $\hat y_{i,f}$. \emph{Always} accepts
every label and serves as the no-withholding reference. \emph{Confidence-only}
(C) uses
\[
s_{i,f}^{(\mathrm C)}
=
\max\{\rho_{i,f},1-\rho_{i,f}\},
\]
where $\rho_{i,f}$ is the upstream stress probability and
$s_{i,f}^{(\mathrm C)}$ is its predicted-class confidence
~\cite{hendrycks2017baseline}. The learned \emph{no-geometry} system (NG)
uses the fixed-label confidence, modality quality/trust, cross-modal, and
train-standardized atypicality evidence defined in Section~\ref{sec:method}.
\emph{ReliaGate} (RG) adds train-reference embedding geometry. NG and RG each
select between logistic and gradient-boosted decision-tree (GBDT) correctness
scorers within every fold. The primary comparison is therefore system-level;
Section~\ref{sec:results-wesad} reports a scorer-matched
43-versus-91-feature sensitivity.

\subsection{Metrics}
Let $F$ be the number of outer folds. In fold $f$, let $n_f$ be the number of test
windows, $\mathcal A_f$ the set accepted by routing, and $y_i$ window $i$'s
protocol-derived reference label. The accepted-error set is
\[
E_f
=
\{i\in\mathcal A_f:\hat y_{i,f}\neq y_i\}.
\]
We report pooled window-level metrics
\begin{equation}
\begin{aligned}
\mathrm{Cov}
&=
\frac{\sum_{f=1}^{F}|\mathcal A_f|}
     {\sum_{f=1}^{F}n_f},
&\qquad
\mathrm{Risk}
&=
\frac{\sum_{f=1}^{F}|E_f|}
     {\sum_{f=1}^{F}|\mathcal A_f|},
\\[3pt]
\mathrm{SF}
&=
\frac{\sum_{f=1}^{F}|E_f|}
     {\sum_{f=1}^{F}n_f},
&\qquad
\mathrm{Reach}
&=
\frac{1}{F}\sum_{f=1}^{F}
\mathbf 1\{|\mathcal A_f|>0\}.
\end{aligned}
\end{equation}
Here $|\cdot|$ denotes set size and $\mathbf 1\{\cdot\}$ is the indicator.
Cov is the accepted-window fraction, Risk is error conditional on acceptance
~\cite{elYaniv2010foundations,traub2024selectiveEvaluation}, SF is the
accepted-and-wrong fraction, and Reach is the fraction of subjects with any
output. When outputs exist, $\mathrm{SF}=\mathrm{Cov}\times\mathrm{Risk}$, so
the metrics must be read jointly.

We also report $N_{\mathrm{acc}}$, accepted macro-F1 and balanced accuracy, and
per-subject coverage/risk; balanced accuracy is reported only when both classes
occur. Subjects with more accepted windows contribute more to pooled values;
per-subject analyses describe output distribution, not demographic fairness
~\cite{jones2021selectiveDisparities,lee2021fairSelective}. Paired 95\% bootstrap
percentile intervals use 10,000 held-out-subject resamples shared by
both systems. Training and selection are not rerun.

\subsection{Subject-Disjoint Validation-Locked Protocol}

Let $u_1,\ldots,u_F$ follow the fixed manifest order. In fold $f$, $u_f$ is
test, $u_{f+1}$ is validation (cyclically), and all other subjects train; every
subject serves once in each held-out role, and test data are reporting-only.
Correctness-scorer training refits only the canonical feature model to obtain
subject-wise OOF predictions; modality networks and trust rows are reused, so
the complete upstream stack is not OOF. Validation/test geometry uses only
outer-training references, whereas selector-training references may retain
other windows from the query subject. Section~\ref{sec:scorers} defines
validation-locked scorer and threshold selection and its limitations. Robustness
checks use $\alpha\in\{0.05,0.10,0.15,0.20\}$, six 60~s-spaced WESAD test
offsets, leave-subject-out references, matched GBDT scorers, and 100 within-fold
geometry permutations. These are descriptive sensitivities, not causal
ablations or risk guarantees.

\paragraph{Code.}
Reproducibility materials are available in the
\href{https://github.com/jadenmoon27/reliagate-wellcomp2026}{ReliaGate repository};
raw signals and record-level artifacts are not redistributed.

%% file: sections/05_results.tex
\section{Results}
\label{sec:results}

Results report held-out-test outcomes at $\alpha=0.10$, an empirical validation
criterion rather than a risk guarantee. We evaluate pooled coverage--error
trade-offs, subject-level availability, and robustness. Window metrics are
pooled; Reach is subject-level; paired 95\% intervals use 10,000 held-out-subject
bootstrap resamples. In Tables~\ref{tab:wesad-decision-quality}--\ref{tab:ubfc-decision-quality},
bolding marks the best thresholded-system point estimate.

\subsection{WESAD Primary Evaluation and Robustness Checks}
\label{sec:results-wesad}

WESAD is the primary system-level comparison. No-geometry (NG) uses 43
features, whereas ReliaGate (RG) adds 48 geometry features. Because NG and RG
select scorers independently, this comparison evaluates complete fold-local
routing systems rather than isolating geometry.

\begin{table}[!htbp]
\caption{WESAD routing at $\alpha=0.10$.
$N_{\mathrm{acc}}$: accepted windows; Conf.: confidence-only; Cov.: coverage;
Risk: accepted risk; SF: silent-failure rate; mF1: accepted macro-F1; BAcc:
accepted balanced accuracy; Reach: subject reach.}
\label{tab:wesad-decision-quality}
\centering
\footnotesize
\setlength{\tabcolsep}{1.6pt}
\begin{tabular*}{\columnwidth}{@{\extracolsep{\fill}}lrrrrrrr@{}}
\toprule
System & $N_{\mathrm{acc}}$ & Cov. & Risk & SF & mF1 & BAcc & Reach \\
\midrule
Always & 2,986 & 1.000 & 0.325 & 0.325 & 0.638 & 0.652 & 1.000 \\
Conf. & 619 & 0.207 & 0.288 & \textbf{0.060} & 0.625 & 0.668 & 0.533 \\
NG & 1,594 & 0.534 & 0.233 & 0.124 & 0.719 & 0.719 &
\textbf{0.933} \\
RG & \textbf{1,908} & \textbf{0.639} & \textbf{0.175} & 0.112 &
\textbf{0.774} & \textbf{0.767} & \textbf{0.933} \\
\bottomrule
\end{tabular*}
\end{table}

At the primary operating point, RG surfaced more labels and made fewer accepted
errors than NG. RG accepted 1,908 windows with 334 errors, whereas NG accepted
1,594 windows with 371 errors: 314 more accepted windows and 37 fewer errors
on different accepted subsets. Both reached 14 of 15 subjects. Confidence-only
had the lowest SF (0.060), but this came with 0.207 coverage and 0.533 Reach. The RG-minus-NG paired difference was $+0.105$ for coverage
(95\% interval [$-0.050$,$+0.265$]) and $-0.058$ for accepted risk
([$-0.154$,$+0.023$]). Both intervals included zero. Descriptively,
non-stress/stress reference-class coverage was $0.538/0.524$ for NG and
$0.660/0.589$ for RG. Error rates among accepted predictions labeled
non-stress/stress were $0.163/0.400$ for NG and $0.132/0.303$ for RG.

Across four $\alpha$ values and six thinned test offsets, RG had
higher-coverage and lower-risk point estimates.
With the scorer fixed, 100 joint within-fold permutations of the 48 geometry
features increased mean area under the risk--coverage curve (AURC; lower is
better) by 0.115 (95\% interval [0.068,0.168]), with positive changes in all
15 folds. This is consistent with ranking sensitivity to geometry, not a causal
effect.

Subject-level summaries were mixed: median coverage was 0.601 for NG and 0.749
for RG, and 11/15 versus 13/15 subjects reached 25\% coverage. However, among
subjects with any output, median accepted risk was 0.107 for NG and 0.163 for
RG. With leave-subject-out
geometry references, RG had coverage/risk $0.654/0.177$ versus primary NG's
$0.534/0.233$. The post-hoc matched gradient-boosted decision-tree (GBDT)
comparison yielded RG $0.679/0.159$ versus NG $0.537/0.243$. Paired coverage
and risk intervals included zero for both sensitivities.

\subsection{UBFC-Phys Additional Controlled Evaluation}
\label{sec:results-ubfc}

\begin{table}[!htbp]
\caption{UBFC-Phys routing at $\alpha=0.10$;
abbreviations follow Table~\ref{tab:wesad-decision-quality}.}
\label{tab:ubfc-decision-quality}
\centering
\footnotesize
\setlength{\tabcolsep}{1.6pt}
\begin{tabular*}{\columnwidth}{@{\extracolsep{\fill}}lrrrrrrr@{}}
\toprule
System & $N_{\mathrm{acc}}$ & Cov. & Risk & SF & mF1 & BAcc & Reach \\
\midrule
Always & 1,848 & 1.000 & 0.366 & 0.366 & 0.607 & 0.614 & 1.000 \\
Conf. & 534 & 0.289 & 0.324 & \textbf{0.094} & 0.649 & 0.652 & 0.554 \\
NG & 1,057 & 0.572 & 0.311 & 0.178 & 0.614 & 0.614 & 0.893 \\
RG & \textbf{1,244} & \textbf{0.673} & \textbf{0.256} & 0.172 &
\textbf{0.696} & \textbf{0.691} & \textbf{0.946} \\
\bottomrule
\end{tabular*}
\end{table}

On UBFC-Phys, RG again accepted more windows while making fewer accepted
errors. RG accepted 1,244 windows with 318 errors, whereas NG accepted 1,057
windows with 329 errors: 187 more windows and 11 fewer errors on different
accepted subsets. The coverage difference was $+0.101$ (paired 95\% interval
[$+0.030$,$+0.175$]) and the accepted-risk difference was $-0.056$
([$-0.105$,$-0.009$]); both intervals excluded zero in the favorable direction.
SF was 0.172 versus 0.178, but its interval ([$-0.047$,$+0.034$]) included
zero.

The same point-estimate direction appeared at all four $\alpha$ values.
Geometry permutation increased mean AURC by 0.058
(95\% interval [0.034,0.083]), with positive changes in 45 of 56 folds. Median
subject coverage was 0.758 for RG and 0.667 for NG. Zero-coverage subjects
decreased from six to three.

\subsection{Within-Dataset E4 Protocol Checks}
\label{sec:results-external}

\begin{table}[!htbp]
\caption{Independent within-dataset E4 outcomes at $\alpha=0.10$;
abbreviations follow Table~\ref{tab:wesad-decision-quality}.}
\label{tab:external-tradeoffs}
\centering
\footnotesize
\setlength{\tabcolsep}{1.5pt}
\begin{tabular*}{\columnwidth}{@{\extracolsep{\fill}}lrrrrrr@{}}
\toprule
Dataset/system & Cov. & Risk & SF & mF1 & BAcc & Reach \\
\midrule
Empatica--NG & 0.681 & 0.178 & 0.121 & 0.731 & 0.756 & 1.000 \\
Empatica--RG & 0.742 & 0.175 & 0.130 & 0.757 & 0.788 & 1.000 \\
PhysioNet--NG & 0.536 & 0.254 & 0.136 & 0.554 & 0.562 & 0.912 \\
PhysioNet--RG & 0.545 & 0.266 & 0.145 & 0.566 & 0.574 & 0.882 \\
\bottomrule
\end{tabular*}
\end{table}

The E4 results were mixed rather than consistently favorable. On
EmpaticaE4Stress, RG had higher coverage, mF1, and BAcc, lower accepted risk,
unchanged Reach, and higher SF. On PhysioNetE4, RG had higher coverage, mF1,
and BAcc, but also higher risk and SF and lower Reach. Paired coverage and risk intervals included zero for both datasets.

%% file: sections/06_discussion.tex
\section{Discussion, Limitations, and Ethics}
\label{sec:discussion}

\noindent\textbf{Interpretation and implications.}
WESAD illustrates why selective wearable systems cannot be judged by accepted
risk alone: pooled coverage and risk favored ReliaGate, but median subject risk
rose despite higher median coverage. This matters for wearable well-being
because surfaced-label error and output availability can move differently
across people. Only UBFC-Phys had favorable primary coverage/risk intervals
excluding zero; E4 outcomes were mixed. The evidence supports ReliaGate as a framework for studying
coverage, surfaced-label error, and subject-level access, not as a universal
geometry improvement. Because scorers differed, the primary comparison is
system-level; matched-scorer intervals included zero, and geometry permutation
was consistent with ranking sensitivity rather than causality.

\noindent\textbf{Limitations.}
Protocol labels are laboratory proxies rather than clinical or free-living
stress, and datasets are not replications or transfer tests. The upstream
classifier omits signed modality posteriors, so conclusions concern routing
around this substrate. One validation subject supports checkpoint, temperature,
scorer, and threshold selection; the WESAD RG GBDT also uses row-level early
stopping. Test subjects remain isolated, but validation reuse and overlapping
windows may yield optimistic selection~\cite{kapoor2023leakage}. Bootstrap
intervals omit selection uncertainty, and thinning only subsamples test rows.
Primary training geometry may retain same-subject windows and encoder exposure;
leave-subject-out removes references only. The 0.10 criterion is empirical, not
a safety guarantee.

\noindent\textbf{Ethics and future work.}
We did not evaluate free-living use, user interpretation, demographic groups,
or sensor-quality differences; per-subject reporting is not a fairness analysis.
Future studies should follow data-use terms, minimize retention, and restrict
access. The present evidence does not support clinical or other high-stakes use.
Longitudinal work should report risk with coverage
~\cite{traub2024selectiveEvaluation} and examine no-output periods, user
understanding, routing drift, and subgroup coverage or error where data permit
~\cite{jones2021selectiveDisparities,lee2021fairSelective}.

%% file: sections/07_conclusion.tex
\section{Conclusion}
\label{sec:conclusion}

ReliaGate frames low-stakes wearable stress inference as fixed-label routing:
after a system predicts a protocol-derived stress/non-stress label, a separate
gate decides whether to surface that unchanged label. Using wearable
reliability cues and train-fitted geometry, ReliaGate evaluates output
availability alongside surfaced-label error. Across four within-dataset
evaluations, the results show favorable but not uniform evidence for
geometry-augmented routing and highlight that pooled risk--coverage gains can
coexist with uneven subject-level outcomes. The contribution is a transparent
framework for studying when wearable predictions should be made available, not
a claim of clinical validity, deployment readiness, causal geometry effects, or
guaranteed risk control. The next empirical step is to test whether this audit
can maintain both acceptable surfaced-label error and adequate per-user
availability under longitudinal, free-living conditions.